\documentclass[preprint,5p,times,twocolumn]{elsarticle}
\usepackage{graphicx}
\usepackage{amsthm}
\usepackage{amsmath}
\usepackage{lineno}
\usepackage{url}
\usepackage{flushend}
\usepackage[latin1]{inputenc}
\usepackage{colortbl}
\usepackage{soul}
\usepackage{multirow}
\usepackage{pifont}
\usepackage{color}
\usepackage{alltt}
\usepackage[hidelinks]{hyperref}
\usepackage{enumerate}
\usepackage{siunitx}
\usepackage{breakurl}
\usepackage{epstopdf}
\usepackage{pbox}
\usepackage{caption}
\usepackage{appendix}
\begin{document}
\begin{frontmatter}

\title{Pixel and feature level based domain adaptation for object detection in autonomous driving}
\author{Yuhu Shan}
\author{Wen Feng Lu}
\author{Chee Meng Chew}

\begin{abstract}
Annotating large-scale datasets to train modern convolutional neural networks is prohibitively expensive and time-consuming for many real tasks. One alternative is to train the model on labeled synthetic datasets and apply it in the real scenes. However, this straightforward method often fails to generalize well mainly due to the domain bias between the synthetic and real datasets. Many unsupervised domain adaptation (UDA) methods were introduced to address this problem but most of them only focused on the simple classification task. This paper presents a novel UDA model which integrates both image and feature level based adaptations to solve the cross-domain object detection problem. We employ objectives of the generative adversarial network and the cycle consistency loss for image translation. Furthermore, region proposal based feature adversarial training and classification are proposed to further minimize the domain shifts and preserve the semantics of the target objects. Extensive experiments are conducted on several different adaptation scenarios, and the results demonstrate the robustness and superiority of the proposed method.
\end{abstract}


\end{frontmatter}

\section{Introduction}
\begin{figure}[!t]\centering
	\includegraphics[width=8.8cm,height=5cm]{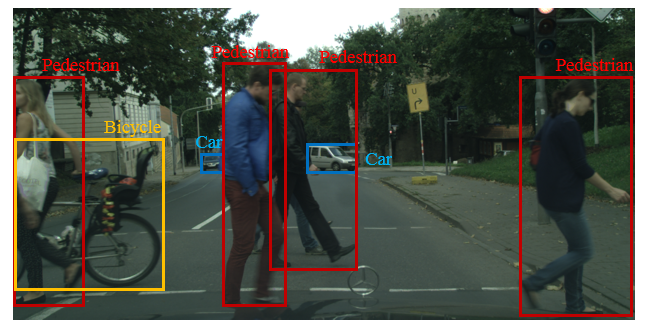}
	\caption{Detection task in autonomous driving system.}
	\label{fig1}
\end{figure}

Object detection aims to assign each object a bounding box along with class label, e.g., ``pedestrian", ``bicycle", ``motorcycle" or ``car" in an image. It plays an important role in modern autonomous driving systems since it is crucial to detect other traffic participants as shown in Fig. \ref{fig1}. Despite the performance of object detection algorithms has been greatly improved since the introduction of AlexNet \cite{alexnet} in 2012, it is still far from satisfactory when it comes to the practical applications, mainly due to the limited data and expensive labeling cost. Supervised learning algorithms based on deep neural networks require large number of fine labeled images, which are extremely difficult to acquire in real cases. For example, it takes almost ninety minutes to annotate one image from the Cityscapes dataset \cite{cityscapes} for driving scene understanding. Even it is already one of the largest driving scene datasets, there are only 2975 training images with fine labels. One promising method to address this problem is to train models on synthetic datasets. Fortunately, with the great progress achieved in graphics and simulation infrastructure, many large-scale datasets with high quality annotations have been produced to simulate different real scenes. However, models trained purely on the rendered images cannot be generalized to real ones, because of the domain shift \cite{domain1,domain2} problem. 

During the past several years, many researchers have proposed various unsupervised domain adaptation (UDA) methods \cite{uda,jiang2019stacked,jiang2018knowledge,guo2017joint} to solve this problem. However, most of them only focused on the simple classification task, which is not suitable for more complex vision tasks such as object detection or semantic segmentation. 
In this paper, we present a new UDA model with adaptations both in pixel and feature spaces to deal with the complex object detection problem in autonomous driving. In the setting of UDA, we generalize the model trained on source dataset with ground truth labels to target dataset without any annotations. 
Perhaps the most similar works to the proposed work are \cite{pixelDA, cycada}. In \cite{pixelDA}, its image generation model is trained based on generative adversarial network (GAN), which is simultaneously combined with the task model for object classification and pose estimation. In \cite{cycada}, the authors attempted to solve the cross-domain semantic segmentation problem based on CycleGAN \cite{cyclegan} and traditional segmentation networks. However, this method needs to train an extra segmentation network to preserve the semantics of translated images, which slows down the whole training process. Actually, it is not necessary to pay the same attention to all the image pixels for some specific tasks like object detection. Target objects, such as ``car" or ``pedestrian" are more important than other objects or stuffs like ``building" or ``sky". Therefore, region proposal based feature adversarial training and classification are proposed in this paper to further minimize the domain shifts and preserve the semantics of the target objects. The developed pixel and feature level based domain adaptation modules can be integrated together and trained end-to-end to pursue better detection performance. Qualitative and quantitative results conducted on several datasets show the robustness of the proposed method.
 
\section{Related work}
\subsection{Object detection}
Object detection \cite{rcnn, fastrcnn} as a fundamental problem in computer vision has achieved great progress since 2012 with the development of deep neural networks. Based on Alexnet, many different convolutional neural networks (CNN), such as VGGnet \cite{vgg}, GoogLeNet \cite{googlenet}, ResNet \cite{resnet}, DenseNet \cite{densenet}, etc., were proposed to learn more powerful deep features from data. Object detection algorithms also benefit from these architectures since better features are also helpful for other vision tasks. Apart from the different network architectures, recent CNN-based object detectors can be mainly divided into two categories: single-stage detectors (YOLO \cite{yolo,yolov3} and SSD \cite{ssd,dssd}) and two-stage detectors (Faster R-CNN \cite{fastercnn} and R-FCN \cite{rfcn}, etc.). Single stage detectors directly predict object labels and bounding box coordinates within one image based on the default anchor boxes, which is fast but not accurate enough. In contrast, two stage detectors firstly generate large number of region proposals based on the CNN features, and then recognize the proposals with heavy head. Therefore, it has better performance than single stage detectors, but the detection speed is slower. Recently, many researchers also attempt to train the detector with only weak labels \cite{jie2017deep, tang2018weakly} or a few ground-truth labels \cite{kang2018few} to deal with more practical application scenarios.  
\begin{figure*}[!t]\centering
	\includegraphics[width=17cm]{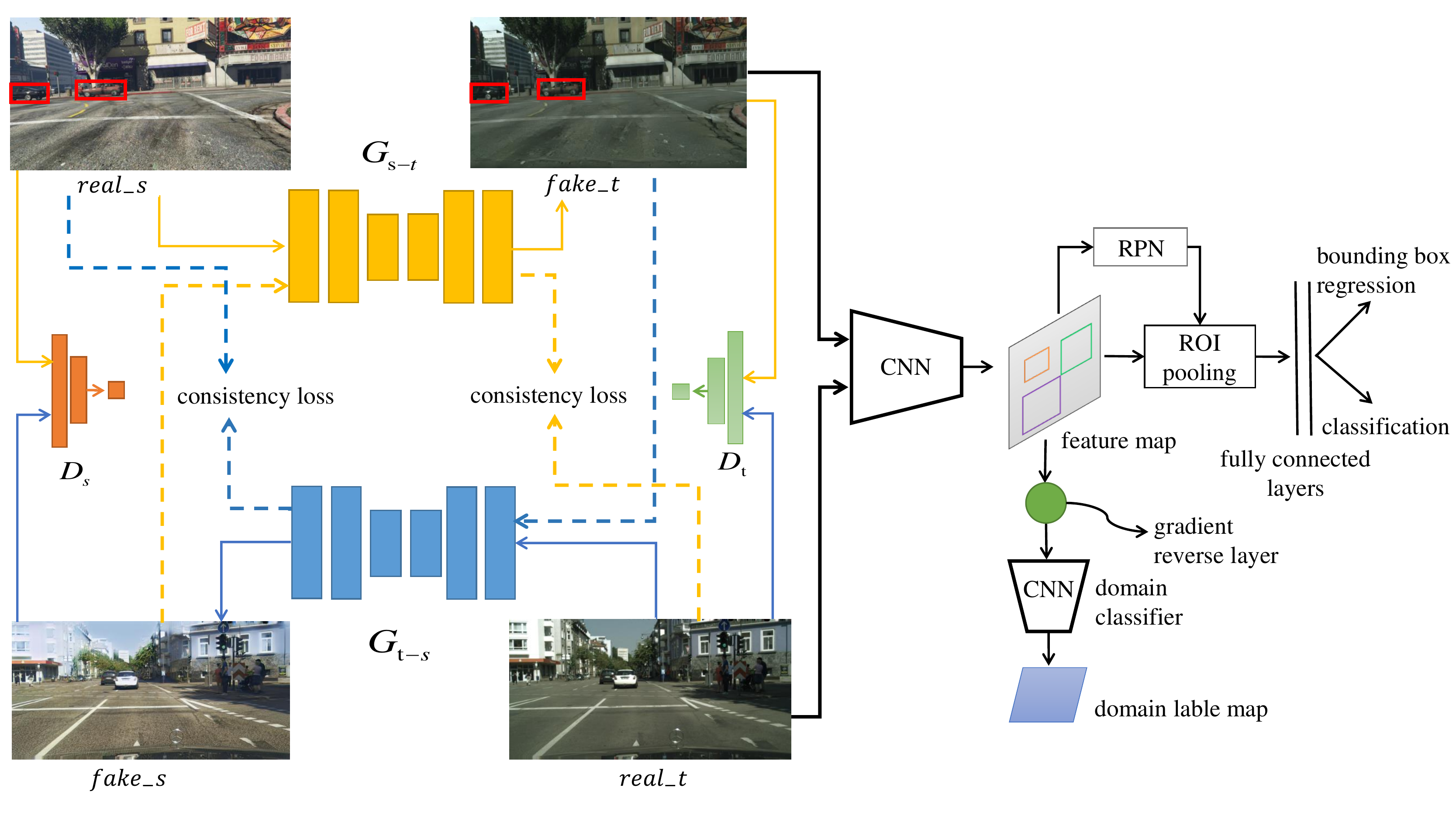}
	\caption{Overall architecture of the proposed learning method. On the left, we show the pixel-level based image translation module, in which source image is firstly converted to the target domain before it is used to train the detection network. On the right, detection network is trained together with the feature-level based adversarial training.}
	\label{fig2}
\end{figure*}
\subsection{Unsupervised domain adaptation}
Unsupervised domain adaptation aims to solve the learning problem in a target domain without labels by leveraging training data in a different but related source domain with ground truth labels. Pan et al. \cite{tca} proposed Transfer Component Analysis (TCA), a kernel method based on Maximum Mean Discrepancy (MMD) \cite{mmd}, to learn better feature representation across domains. Based on TCA, \cite{jda} provided a new method of Joint Distribution Adaptation to jointly adapt both the marginal distribution and the conditional distribution, which was robust for substantial distribution difference. Jiang et al. \cite{jiang2017theoretic} proposed to close the second moments of the source and target domain distributions to obtain better adaptation performance. Recently, with the advent of deep learning, many works were proposed to learn deep domain invariant features within the neural networks. For example, Long et al. \cite{long-adaptation, long-adaptation-residual} proposed to embed hidden network features in a reproducing kernel Hilbert space and explicitly measure the difference between the two domains with MMD and its variants. Sun et al. \cite{sun} tried to minimize domain shift by matching the second order statistics of feature distributions between the source and the target domains. Rather than explicitly modeling the term to measure the domain discrepancy, another stream of works utilized adversarial training to implicitly find the domain invariant feature representations. Ganin et al. \cite{dann, dann1} added one more domain classifier to the deep neural network model to classify the domains of the inputs. Adversarial training is conducted through reversing the gradients from the domain classification loss. Rather than using shared feature layers, Tzeng et al. \cite{adda} proposed to learn indistinguishable features for the target domain data by training a separate network with objectives similar to the traditional GAN model \cite{gan}. Then it is combined with the classifier trained on the source domain for recognition tasks. In \cite{dsn}, the authors argued that each domain should have its own specific features and only part of the features were shared between the different domains. Therefore, they explicitly modeled the private and shared domain representations and only conducted adversarial training on these share ones. 

Despite many methods have been proposed to solve the UDA problem, most of them only focus on the simple classification task. Very limited works were conducted related to more complex tasks such as object detection or semantic segmentation. As far as we know, \cite{yuhua} is the first work to deal with the domain adaptation problem for object detection. They conducted adversarial training on features from convolutional and fully connected layers, along with the regular training for detection task. A domain label consistency check was used as a regularizer to help the network to learn better domain invariant features. Although good detection results were achieved in \cite{yuhua}, we argue that conducting adaptations on both feature and image pixel spaces would be a better alternative since adapting high-level features can fail to model the low-level image details. Therefore, the proposed work is also closely related to image translation.

\subsection{Image-to-image translation}
Currently, many works have been done to convert images into another style. \cite{pair1,pair2,pair3,pixel2pixel,bicyclegan} conducted image translation based on the assumption of available paired training images, which is not fit to the problem of unsupervised domain adaptation. Several other recent works tried to solve the problem with unpaired images. \cite{couplegan,crossmodal} shared part of the network weights to learn the joint distribution of multi-modal images. PixelDA \cite{pixelDA} proposed a novel GAN based architecture to convert images across domains. However, this method needs prior knowledge about which parts of the image contributing to the content similarity loss. Neural style transfer \cite{perceptual,hangzhangstyle,gatys1} is another kind of method to convert one image into another style while preserving its own contents through optimizing the pixel values during back-propagation process. One shortcoming of style transfer is that it only targets on translation between two specific images while not at the dataset level. Recently proposed CycleGAN model \cite{cyclegan} is a promising method for unpaired image translation. The authors utilized cycle consistency loss to regularize the generative model and hence to preserve structural information of the transferred image. However, this method can only guarantee that, one area if occupied by one object before the translation, will also get occupied after the generation process. Semantics of the pixels are not guaranteed to be consistent with this only cycle consistence loss.      

\section{Method}
We tackle the problem of unsupervised cross-domain object detection with the assumption of available source images $X_s$ with ground truth labels $Y_s$ and target images $X_t$ without any labels. Our target is to train the detection network with source dataset, which also need to perform well on the target dataset. The whole framework is shown in Fig. \ref{fig2}. Our model consists of two modules: 1) pixel-level domain adaptation (PDA) mainly based on CycleGAN, and 2) feature-level domain adaptation (FDA) based on Faster R-CNN. The two modules can be integrated together and trained in an end-to-end way to pursue better performance. Source images are firstly converted into the target image style. Then the transformed images are used to train the object detector and domain classifier, along with the sampled images from the target dataset. 

\subsection{Pixel-level based domain adaptation}
We firstly introduce the pixel-level based domain adaptation module. As shown in Fig. \ref{fig2}, two symmetric generative networks $G_{s-t}$ and $G_{t-s}$ are employed to generate images $fake_{-}t$ and $fake_{-}s$ separately in two domains. Another two discriminators $D_{s}$ and $D_{t}$ are trained to distinguish the real sampled and fake generated images. The whole training process runs in a min-max manner, in which the generators always try to generate images which cannot be distinguished from the real ones. The discriminators are trained simultaneously to be good at classifying real and fake images. The whole objectives of generator $G_{s-t}$ and discriminator $D_{t}$ can be formulated as Eq. (\ref{eq1}).
\begin{equation}
\begin{aligned}
&L_{GAN}(D_{t},G_{s-t})= \mathbb{E}_{t\sim X(t)}[logD_{t}(x)] \\ 
&+\mathbb{E}_{s\sim X(s)}[log(1-D_{t}(G_{s-t}(s)))].
\end{aligned}  
\label{eq1}
\end{equation}

Similar equation can also be formulated for $G_{t-s}$ and $D_{s}$ as $L_{GAN}(D_{s},G_{t-s})$, which is ignored here. This kind of GAN objectives can, in theory, learn the mapping functions $G_{s-t}$ and $G_{t-s}$ to produce images identically sampled from the data distribution of $X_{t}$ and $X_{s}$. However, it also faces the problem of mode collapse \cite{gan} and losing the structural information of the source images. To address these problems, cycle consistency loss is adopted here to force the image $fake_{-}t$ generated by $G_{s-t}$ to have identical result as $real_{-}s$ after it is sent into generator $G_{t-s}$ and vice versa. The whole cycle consistency loss is formulated in Eq. (\ref{eq2}),
\begin{equation}
\begin{aligned}
&L_{cyc}(G_{t-s},G_{s-t})= \mathbb{E}_{t\sim X(t)}[||G_{s-t}(G_{t-s}(x_{t}))-x_{t}||_{1}] \\  
&+\mathbb{E}_{s\sim X(s)}[||G_{t-s}(G_{s-t}(x_{s}))-x_{s}||_{1}].
\end{aligned}
\label{eq2}
\end{equation}
Therefore, the full objective for CycleGAN training is 
\begin{equation}
\begin{aligned}
&L_{cyc-gan}(G_{t-s},G_{s-t},D_t,D_s)= L_{GAN}(D_t,G_{s-t}) + \\ &L_{GAN}(D_s,G_{t-s})+\lambda_{cyc}L_{cyc}(G_{t-s},G_{s-t}), 
\end{aligned}
\label{eq3}
\end{equation}
with the target of solving
\begin{equation}
\begin{aligned}
&G_{s-t}^*, G_{t-s}^*= \arg{\min_{G_{s-t},G_{t-s}}}\max_{D_s,D_t}L_{cyc-gan}(G_{t-s},G_{s-t},D_t,D_s).
\end{aligned}
\label{eq4}
\end{equation}

\subsection{Feature-level based domain adaptation}
Our detection network is based on the famous framework of Faster R-CNN. Specifically, region proposal network (RPN) is trained to generate region proposals and Fast R-CNN \cite{fastrcnn} trained for bounding box classification and regression. A small fully convolutional network is newly added to the framework for further domain adversarial training. Specifically, the inputs are features extracted from the final convolutional features by the corresponding region proposals. Gradients generated by the detection and reversed domain classification losses will flow into the shared convolution layers to learn domain-invariant features for object detection.
\paragraph{Losses for Faster R-CNN}
Assuming there are $m$ categories in the detection task, the region classification layer will output $m+1$ dimension probability distribution for each region proposal, $p_{obj}=(p_{obj}^{0},p_{obj}^{1},...,p_{obj}^{m})$, with one more category for the background. $p_{obj}^*$ is used to represent the ground truth label for the region proposal. Box coordinate $t_{obj}=(t_x,t_y,t_w,t_h)$ is predicted for each possible class by the bounding box regression layer to approach the ground truth regression target $t_{obj}^*$. Here $t_{obj}$ and $t_{obj}^*$ are normalized as \cite{fastercnn} for better training of the network. Similarly, we use $p_{rp}$ and $p_{rp}^*$ for the training of RPN. Since RPN is only trained to discriminate object/non-object, labels $p_{rp}^*$ can only be 1 or 0. The full objectives for Faster R-CNN training can be formulated as Eq. \eqref{eq5}, 
\begin{equation}
\begin{aligned}
&L_{det}(p,p^*,t,t^*) = L_{cls-det}(p_{obj},p^*_{obj})+[p_{obj}^* \ge 1] \\
&L_{loc-det}(t_{obj},t^*_{obj}) + L_{cls-rpn}(p_{rp},p^*_{rp}) \\
&+[p_{rp}^* \ge 1]L_{loc-rpn}(t_{rp},t^*_{rp}),
\end{aligned}
\label{eq5}
\end{equation}
in which $L_{cls-det}$ and $L_{cls-rpn}$ indicate the cross-entropy loss for classification. $L_{loc-det}$ and $L_{loc-rpn}$ represent the smooth $L_{1}$ loss \cite{fastrcnn} for bounding box regression. The Iverson bracket indicator function $[p^* \ge 1]$ evaluates to 1 when the true object category $p^* \ge 1$ and 0 otherwise.

\paragraph{Losses for FDA training}
As shown in Fig. \ref{fig2}, domain classifier learns to classify $fake_{-}t$ as label $d=0$ and $real_{-}t$ as label $d=1$. Then the gradients are reversed before they flow into the shared convolutional layers. Loss of the domain adversarial training is shown in Eq. \eqref{eq6},
\begin{equation}
L_{domain} = - \sum_{i,j}[dlogp_{i,j}+(1-d)log(1-p_{i,j})],
\label{eq6}
\end{equation}
in which $p_{i,j}$ indicates the classification output on the $\emph{i-th}$ region proposal of the $\emph{j-th}$ image.
Based on the above equations, we can formulate the full training objectives in Eq. \eqref{eq7}, where $\lambda_{1},\lambda_{2}$ are the weights to balance the different losses,
\begin{equation}
\begin{aligned}
L_{full} = L_{det} + \lambda_{1} L_{domain}+\lambda_{2}L_{cyc-gan}. 
\end{aligned}
\label{eq7}
\end{equation}

\section{Implementation}
\subsection{Datasets}
\begin{figure*}[!t]\centering
	\includegraphics[width=14cm]{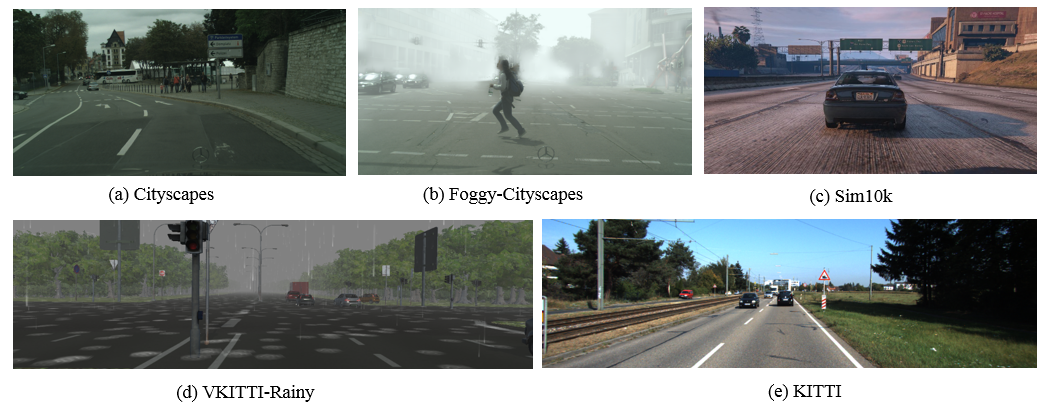}
	\caption{Sampled images from the five datasets to show the domain bias.}
	\label{fig3}
\end{figure*}

To test the validity of the proposed method and also compare with current state-of-the-art (SOTA) work \cite{yuhua}, we choose Cityscapes, KITTI \cite{kitti}, Foggy-Cityscapes \cite{foggy},  VKITTI-Rainy \cite{vkitti} and Sim10k \cite{sim10k} datasets for the experiments. Cityscapes dataset has 2975 training images and 500 images for validation. Eight classes of common traffic participants are annotated with instance labels. KITTI is another famous dataset for benchmarking different vision tasks in autonomous driving. There are 7481 labeled training images with bounding boxes for categories ``car", ``pedestrian" and ``cyclist". Foggy-Cityscapes, VKITTI-Rainy and Sim10k are synthetic datasets which simulate different driving scenes. Particularly, Foggy-Cityscapes and VKITTI-Rainy are rendered based on the real Cityscapes and KITTI datasets to simulate the foggy and rainy weathers. Sim10k has 10,000 training images which are collected from the computer game of GTA5 and annotated automatically by access to the original game engine. In Sim10k dataset, only objects ``car" are annotated with bounding boxes for the detection task. Therefore, we only calculate and compare the ``car" detection result for the experiments based on Sim10k in this paper. Five exampled images randomly sampled from the above five datasets are shown in Fig. \ref{fig3} to show their domain differences. 

\subsection{Implementation details}
We use U-Net structure \cite{unet} with skip connections between layers for the two generators in the pixel adaptation module and PatchGAN \cite{pixel2pixel} for the other two discriminators. Instance normalization is adopted since it is more effective as stated in original CycleGAN paper. For the detection network, we use VGG16 \cite{vgg} as the backbone and a small fully convolutional network for domain adversarial training. Inputs for the feature-level based adversarial training are the cropped Conv5 features of VGG16 based on the 64 region proposals generated by the RPN module. In our practical training process, we choose to firstly pre-train the detection and image translation networks separately and then conduct the end-to-end training based on these two pre-trained models. This mainly considers the fact that most of the generated images are quite noisy in the start training stage of the pixel-level adaptation module. We train the PDA module with Adam optimizer and an initial learning rate of 0.0002. After 30 epochs, the learning rate linearly decays to be zero in the following training process for another 30 epochs. FDA module is trained together with the object detection network with initial learning rate of 0.001 based on the standard SGD algorithm. After 6 epochs, we reduce the learning rate to 0.0001 and train the network for another 3 epochs. Gradients from the domain classifier are reversed before they flow into the shared CNN layers during the back-propagation. For the end-to-end training, all the above initial learning rates are scaled down by ten times. We then finetune the whole network for another 10 epochs with $\lambda_{1}$ and $\lambda_{2}$ set as 0.5.    
    
\section{Results and discussion}
We show our experimental results under three adaptation scenarios: ``synthetic to real", ``cross different weathers" and ``cross different cameras". Specifically, experiments are conducted on ``Sim10k $\rightarrow$ Cityscapes" and ``Sim10k $\rightarrow$ KITTI" for the scenario of ``synthetic to real", ``Cityscapes $\rightarrow$ Foggy-Cityscapes" and ``KITTI $\rightarrow$ VKITTI-Rainy" for ``cross different weathers", ``Cityscapes $\rightarrow$ KITTI" and ``KITTI $\rightarrow$ Cityscapes" for ``cross different cameras". For each dataset pair, we conduct three experiments with considering the PDA, FDA modules and the final end-to-end training. To test the validity of the PDA module, we only train the network for image translation, and use the translated images to train a pure detection network. For the FDA training, we directly use the source images as inputs for the detection training. Randomly sampled images from the target domain are combined with the source images for the feature-level based adversarial training. Finally, we integrate the two modules together and train the whole network in an end-to-end way. All the results are evaluated with the commonly used metrics of Average Precision (AP) and mean Average Precision (mAP). IoU threshold is set as 0.5.
\begin{table}[htb]
	\centering
	\newcommand{\tabincell}[2]{\begin{tabular}{@{}#1@{}}#2\end{tabular}}
	\caption{Detection results (AP (\%) of ``car") are evaluated on the Cityscapes validation and KITTI training datasets by using Sim10k as the source dataset.}
	\label{table1}
	\renewcommand{\arraystretch}{0.8}
	\small
	\resizebox{\columnwidth}{!}{
		\begin{tabular}{ccc}
			\hline
			&\multirow{2}{*}{\tabincell{c}{Sim10k $\rightarrow$ Cityscapes}}
			& \multirow{2}{*}{\tabincell{c}{Sim10k $\rightarrow$ KITTI}} \\
			& &    \\
			\hline
			\multirow{2}{*}{Faster R-CNN} & \multirow{2}{*}{30.1} & \multirow{2}{*}{52.7}\\
			& & \\
			\multirow{2}{*}{SOTA\cite{yuhua}}& \multirow{2}{*}{39.0} & \multirow{2}{*}{--}  \\
			& & \\
			\multirow{2}{*}{\tabincell{c}{Faster R-CNN w/ PDA}} & \multirow{2}{*}{37.8} &  \multirow{2}{*}{58.4} \\
			& &  \\
			\multirow{2}{*}{\tabincell{c}{Faster R-CNN w/ FDA}} & \multirow{2}{*}{33.8} & \multirow{2}{*}{55.3} \\
			& &   \\
			\multirow{2}{*}{\tabincell{c}{Faster R-CNN w/ (PDA+FDA)}} & \multirow{2}{*}{\textbf{39.6}} & \multirow{2}{*}{\textbf{59.3}} \\
			& &  \\
			\hline
		\end{tabular}
	}
\end{table} 
\subsection{Synthetic to real}
\label{syn-to-real}
In this scenario, object detector is trained on the synthetic dataset and evaluated on the real datasets. 
\paragraph{Sim10k $\rightarrow$ Cityscapes}
To study the efficacy of the proposed method, we firstly consider to remove the domain bias between the computer synthetic dataset of Sim10k and the real dataset of Cityscapes. Table \ref{table1} shows our results with different adaptation modules. Our baseline result with Faster R-CNN is trained purely on the source Sim10k and evaluated on Cityscapes directly. The calculated mAP is 30.1\% with the VGG16 backbone. Compared with the baseline and current SOTA, the proposed method obtains 9.5\% and 0.6\% performance gains, respectively. Specifically, the proposed feature-level based adversarial training can improve the performance of baseline to 33.8\%. The PDA module can bring 7.78\% gains. When an end-to-end training is conducted, we can obtain better results than current SOTA.

\paragraph{Sim10k $\rightarrow$ KITTI}
To further check the proposed method's robustness to the scenario of ``synthetic to real", experiment is conducted by using the KITTI dataset as the target dataset. Since there are no other works reporting the detection results under this specific setting, we only compare our results with the baseline ones. All the results are evaluated on the KITTI training split just like \cite{yuhua}. As shown in Table \ref{table1}, the baseline Faster R-CNN network has an AP of 52.7\%. It can then be improved to 55.3\% and 58.4\% by using the proposed FDA and PDA modules separately. Through conducting the end-to-end training, we can get the highest result of 59.3\%.
\begin{table}[htb]
	\centering
	\newcommand{\tabincell}[2]{\begin{tabular}{@{}#1@{}}#2\end{tabular}}
	\caption{Detection results are evaluated for the adaptations of Cityscapes $\rightarrow$ Foggy-Cityscapes and KITTI $\rightarrow$ VKITTI-Rainy.}
	\label{table2}
	\renewcommand{\arraystretch}{1.0}
	\small
	\resizebox{\columnwidth}{!}{
		\begin{tabular}{ccc}
			\hline
			&\multirow{2}{*}{\tabincell{c}{Cityscapes $\rightarrow$ \\ Foggy-Cityscapes}}
			& \multirow{2}{*}{\tabincell{c}{KITTI $\rightarrow$ \\ VKITTI-Rainy}} \\
			& &    \\
			\hline
			\multirow{2}{*}{Faster R-CNN} & \multirow{2}{*}{18.8} & \multirow{2}{*}{40.0}\\
			& & \\
			\multirow{2}{*}{SOTA\cite{yuhua}}& \multirow{2}{*}{27.6} & \multirow{2}{*}{--}  \\
			& & \\
			\multirow{2}{*}{\tabincell{c}{Faster R-CNN w/ PDA}} & \multirow{2}{*}{27.1} &  \multirow{2}{*}{51.3} \\
			& &  \\
			\multirow{2}{*}{\tabincell{c}{Faster R-CNN w/ FDA}} & \multirow{2}{*}{23.6} & \multirow{2}{*}{44.5} \\
			& &   \\
			\multirow{2}{*}{\tabincell{c}{Faster R-CNN w/ (PDA+FDA)}} & \multirow{2}{*}{\textbf{28.9}} & \multirow{2}{*}{\textbf{52.2}} \\
			& &  \\
			\hline
		\end{tabular}
	}
\end{table} 

\subsection{Cross different weathers}
In addition to the ``synthetic to real" scenario, domain bias caused by different weathers are also considered in the experiments. Specifically, the proposed method is testified to deal with the foggy and rainy weathers in the driving scenes. 
\paragraph{Cityscapes $\rightarrow$ Foggy-Cityscapes}
Cityscapes and its foggy version are used as the source and target datasets in this experiment. All the training settings are the same with Section \ref{syn-to-real}. The main target here is to adapt the detector to fit to the foggy weather by utilizing the labeled images collected in the clear weather. Experimental results are shown in Table \ref{table2}. We can finally achieve 10.1\% and 1.8\% performance gains compared to the baseline result and current SOTA. Both of the two modules can largely improve the detection performance under the foggy circumstance. Specifically, baseline results of Faster R-CNN can be improved from 18.8\% to 23.6\% with FDA and 27.1\% with PDA. 

\paragraph{KITTI $\rightarrow$ VKITTI-Rainy}
To certify the method's robustness to the rainy weather, VKITTI-Rainy is used as the target dataset, and KITTI as the source dataset. Because previous SOTA work does not report their results under this circumstance, experimental results are only compared with the baseline Faster R-CNN. Consistent improvements are achieved in this rainy condition. Detection results can be finally improved from 40.0\% to 52.2\% with the end-to-end training. It also can be seen that PDA module performs much better than the FDA module in this case. 

\begin{table}[htb]
	\centering
	\newcommand{\tabincell}[2]{\begin{tabular}{@{}#1@{}}#2\end{tabular}}
	\caption{Detection results (AP (\%) of ``car") are evaluated for the adaptations of Cityscapes $\rightarrow$ KITTI and KITTI $\rightarrow$ Cityscapes.}
	\label{table3}
	\renewcommand{\arraystretch}{0.8}
	\small
	\resizebox{\columnwidth}{!}{
		\begin{tabular}{ccc}
			\hline
			&\multirow{2}{*}{\tabincell{c}{Cityscapes $\rightarrow$ KITTI}}
			& \multirow{2}{*}{\tabincell{c}{KITTI $\rightarrow$ Cityscapes}} \\
			& &    \\
			\hline
			\multirow{2}{*}{Faster R-CNN} & \multirow{2}{*}{53.5} & \multirow{2}{*}{30.2}\\
			& & \\
			\multirow{2}{*}{SOTA\cite{yuhua}} & \multirow{2}{*}{64.1} &  \multirow{2}{*}{38.5} \\
			& &  \\
			\multirow{2}{*}{\tabincell{c}{Faster R-CNN w/ PDA}} & \multirow{2}{*}{64.4} &  \multirow{2}{*}{41.1} \\
			& &  \\
			\multirow{2}{*}{\tabincell{c}{Faster R-CNN w/ FDA}} & \multirow{2}{*}{58.6} & \multirow{2}{*}{34.5} \\
			& &   \\
			\multirow{2}{*}{\tabincell{c}{Faster R-CNN w/ (PDA+FDA)}} & \multirow{2}{*}{\textbf{65.6}} & \multirow{2}{*}{\textbf{41.8}} \\
			& &  \\
			\hline
		\end{tabular}
	}
\end{table} 
\subsection{Cross different cameras}
To fully compare with current SOTA work, experiments are further conducted to deal with the domain bias caused by using different cameras. Two real datasets KITTI and Cityscapes are selected as the source and target datasets alternatively. The results are shown in Table \ref{table3}. With Cityscapes as the source dataset, baseline results of Faster R-CNN can be improved from 53.5\% to 64.4\% (with PDA) and 58.6\% (with FDA). When KITTI is used as the source dataset, the detection results can be improved from 30.2\% to 41.1\% (with PDA) and 34.5\% (with FDA). The proposed method achieves better performance than current SOTA in both adaptation settings, with improvements of 1.5\% and 2.3\%, respectively. 

\subsection{Analysis and discussion}
\begin{table}[htb]
	\centering
	\newcommand{\tabincell}[2]{\begin{tabular}{@{}#1@{}}#2\end{tabular}}
	\caption{Quantitative ablation mAP results (\%) of the different domain adaptation scenarios.}
	\label{table4}
	\renewcommand{\arraystretch}{1.2}
	\small
	\resizebox{\columnwidth}{!}{
		\begin{tabular}{ccccc}
			\hline
			&\multirow{2}{*}{\tabincell{c}{Faster R-CNN}}
			& \multirow{2}{*}{\tabincell{c}{Faster R-CNN \\ w/ PDA}} & \multirow{2}{*}{\tabincell{c}{Faster R-CNN \\ w/ FDA}} & \multirow{2}{*}{\tabincell{c}{Faster R-CNN \\ w/ PDA+FDA}} \\
			& & &&   \\
			\hline
			\multirow{2}{*}{\tabincell{c}{Sim10k \\ $\rightarrow$ Cityscapes}} & \multirow{2}{*}{30.1} & \multirow{2}{*}{\tabincell{c}{37.8 (+7.7)}} & \multirow{2}{*}{\tabincell{c}{33.8 (+3.7)}} & \multirow{2}{*}{\tabincell{c}{39.6 (+9.5)}}\\
			& & && \\
			\multirow{2}{*}{\tabincell{c}{Sim10k \\ $\rightarrow$ KITTI}} & \multirow{2}{*}{\tabincell{c}{52.7}} &  \multirow{2}{*}{\tabincell{c}{58.4 (+5.7)}} &  \multirow{2}{*}{\tabincell{c}{55.3 (+2.6)}} &  \multirow{2}{*}{\tabincell{c}{59.3 (+6.6)}} \\
			& & && \\
			\multirow{2}{*}{\tabincell{c}{Cityscapes \\ $\rightarrow$ Foggy-Cityscapes}} & \multirow{2}{*}{\tabincell{c}{18.8}} & \multirow{2}{*}{\tabincell{c}{27.1 (+8.3)}} &  \multirow{2}{*}{\tabincell{c}{23.6 (+4.8)}} &  \multirow{2}{*}{\tabincell{c}{28.9 (+10.1)}} \\
			& & && \\
			\multirow{2}{*}{\tabincell{c}{Cityscapes \\ $\rightarrow$ KITTI}} & \multirow{2}{*}{\tabincell{c}{53.5}} & \multirow{2}{*}{\tabincell{c}{64.4 (+10.9)}} &  \multirow{2}{*}{\tabincell{c}{58.6 (+5.1)}} &  \multirow{2}{*}{\tabincell{c}{65.6 (+12.1)}} \\
			& &   &&\\
			\multirow{2}{*}{\tabincell{c}{KITTI \\ $\rightarrow$ Cityscapes}} & \multirow{2}{*}{\tabincell{c}{30.2}} & \multirow{2}{*}{\tabincell{c}{41.1 (+10.9)}} &  \multirow{2}{*}{\tabincell{c}{34.5 (+4.3)}} &  \multirow{2}{*}{\tabincell{c}{41.8 (+11.6)}} \\
			& &  &&\\
			\multirow{2}{*}{\tabincell{c}{KITTI \\ $\rightarrow$ VKITTI-Rainy}} & \multirow{2}{*}{\tabincell{c}{40.0}} & \multirow{2}{*}{\tabincell{c}{51.3 (+11.3)}} &  \multirow{2}{*}{\tabincell{c}{44.5 (+4.5)}} &  \multirow{2}{*}{\tabincell{c}{52.2 (+12.2)}} \\
			& &  &&\\
			\hline
		\end{tabular}
	}
\end{table} 
\paragraph{PDA versus FDA} Two modules (PDA and FDA) are proposed to reduce the domain bias between the source and target datasets. A final end-to-end training of the whole network is also conducted to pursue better performance of cross-domain object detection. The quantitative ablation results are further summarized in Table \ref{table4} to compare the effectiveness of the different modules.
\begin{figure}[htb]
	\begin{center}
		\includegraphics[width=8cm, height=5cm]{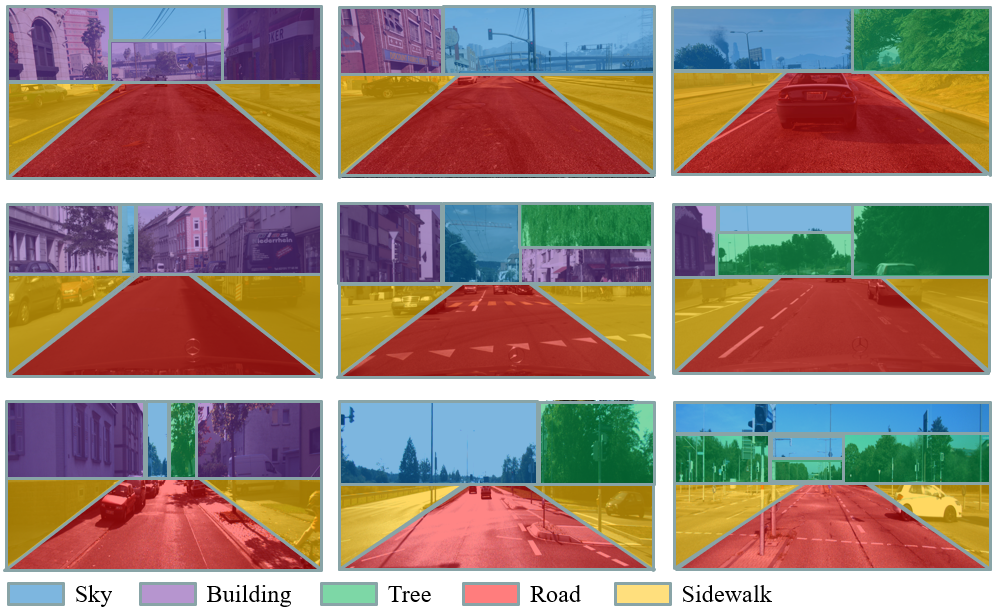}
	\end{center}
	\vspace{-0.2cm}
	\caption{Overall structure layouts of the different driving scenes. From up to bottom, we show the images from Sim10k, Cityscapes and KITTI, respectively. Different driving scenes share the similar layouts along the ``road" (\emph{i.e.} the red and yellow areas.)}
	\label{fig4}
\end{figure}

From Table \ref{table4}, it can be seen that both of the two modules work effectively to improve the detector's performance under different domain bias. Image pixel-level transformation performs much better than the feature-level based adversarial training, with an average two times of improvements. Since the PDA module is based on CycleGAN, semantics of the image pixels are not guaranteed to be consistent after the translation process. However, the experimental results show that the image translation still works effectively to improve the detector's performance on the target datasets. Two reasons can be explained to this phenomenon. The first one is that the translated images, despite not perfect, can help the detector to learn generic target-oriented CNN features. The second one is that traffic participants are the main concerns of the object detection task. Semantics of these objects can be mostly kept with PDA since the structure layouts along the ``road" are similar among different driving scenes as shown by Fig. \ref{fig4}. Despite the upper parts of the images (\emph{i.e.} the purple, blue and green areas) vary across different scenes, the lower parts (\emph{i.e.} the red and yellow areas) are quite similar. In these similar areas, CycleGAN based PDA would mainly focus on the translation of color and textures. 

\begin{figure}[htb]
	\begin{center}
		\includegraphics[width=8cm, height=10cm]{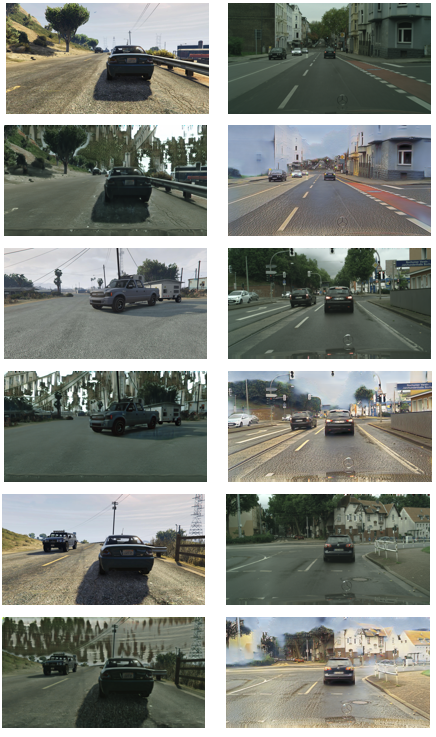}
	\end{center}
	\vspace{-0.2cm}
	\caption{Qualitative image translation results between Sim10k and Cityscapes. On the left, we show the translations from Sim10k to Cityscapes. Translation results from Cityscapes to Sim10k are shown on the right.}
	\label{fig5}
\end{figure}

Fig. \ref{fig5} shows more image translation results between Sim10k and Cityscapes to illustrate this phenomenon. Generally, the synthetic dataset of Sim10k owns quite different structure layouts with the real Cityscapes dataset. Most of the images in Sim10k are taken from the scene of high-way, while Cityscapes are mainly taken from the cities. Therefore, it can be seen from Fig. \ref{fig5} that many image pixels of Sim10k are mapped into ``tree" or ``building" to fit to the structure layout of Cityscapes. Similarly, pixels of ``tree" and ``building" are largely mapped into ``sky" in the translation from Cityscapes to Sim10k. Nevertheless, objects along the ``road" can still be maintained since similar layouts are shared between the two datasets. The translations are mainly focused on color and textures as shown in Fig. \ref{fig5}.

\paragraph{With/without the end-to-end training}
\begin{figure}[htb]
	\begin{center}
		\includegraphics[width=8cm, height=8cm]{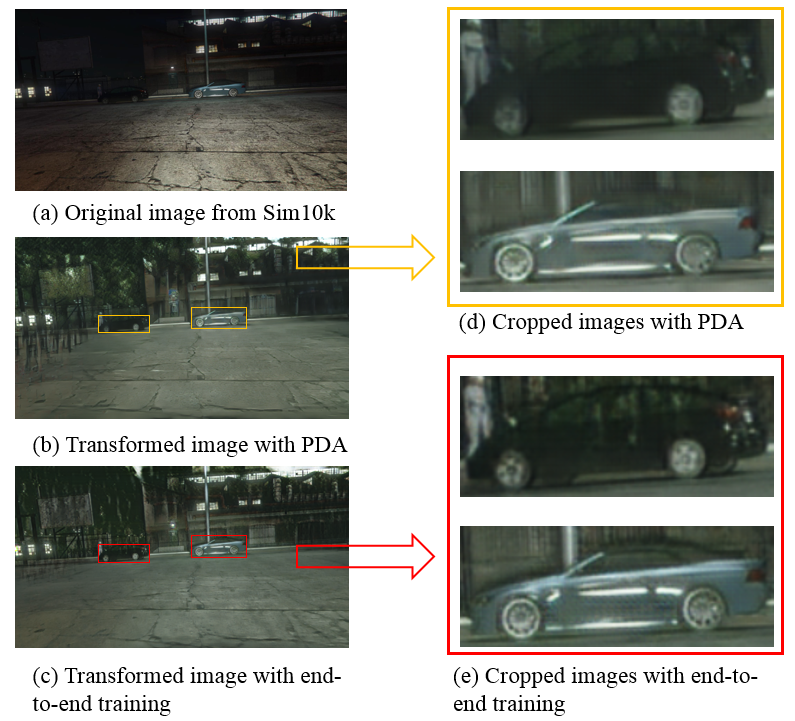}
	\end{center}
	\vspace{-0.2cm}
	\caption{Image translation results are shown with PDA (the yellow rectangular box) and the end-to-end training (the red rectangular box), respectively.}
	\label{fig6}
\end{figure}
During the experiments, an end-to-end training of the whole network is further conducted for another 10 epochs. In this training stage, more attentions are paid to the target objects. PDA module would also attempt to generate more realistic objects to reduce the adversarial training loss of FDA. Quantitatively, by conducting the end-to-end training, more improvements around 1\% $\thicksim$ 2\% can be further achieved for the detection task. Qualitative results are also shown in Fig. \ref{fig6} for better comparison of the translated images. Results with only PDA are shown in the yellow rectangular boxes and the end-to-end ones in the red rectangular boxes. It can be seen that better results are generated after the end-to-end training. When the target objects are close to the background or out of the share structured layouts, one potential problem is that the objects may get merged with the background or falsely translated into other semantics. With the end-to-end training, more details of the target objects can be maintained, such as the roof of the car. 

\begin{figure}[htb]
	\begin{center}
		\includegraphics[width=8.5cm, height=8cm]{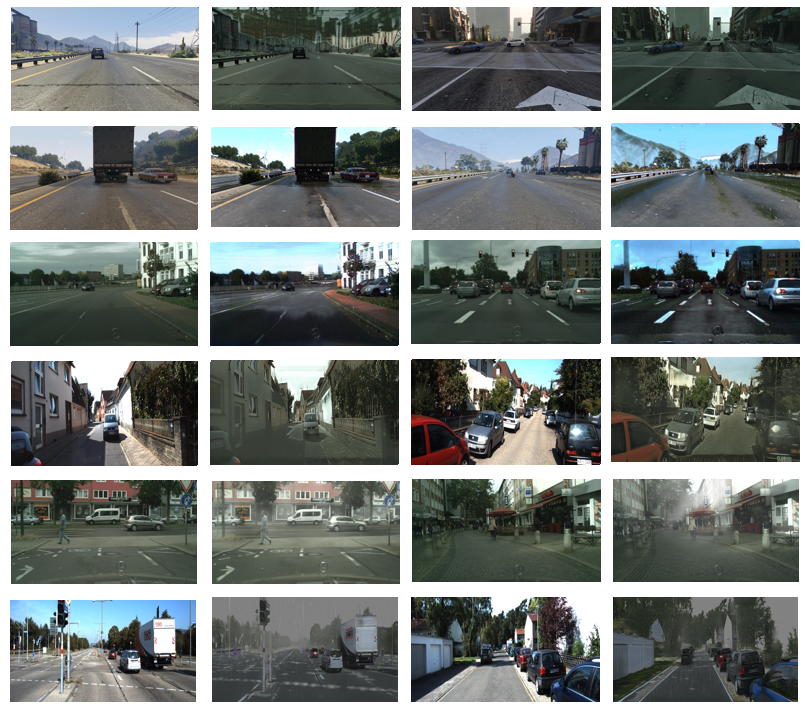}
	\end{center}
	\caption{Qualitative results of the final translated images. From up to bottom, we show the image translations of Sim10k $\rightarrow$ Cityscapes, Cityscapes $\rightarrow$ Foggy-Cityscapes, Sim10k $\rightarrow$ KITTI, Cityscapes $\rightarrow$ KITTI, KITTI $\rightarrow$ Cityscapes and KITTI $\rightarrow$ VKITTI-Rainy. From left to right, we show the source and translated target domain images, alternatively.}
	\label{fig7}
\end{figure}

\paragraph{Analysis of the qualitative results}
Fig. \ref{fig7} shows the qualitative image translation results under different scenarios. Row 1 and 2 show the translations between synthetic images of Sim10k and realistic ones of Cityscapes and KITTI. In this ``synthetic to real" scenario, structures of the images are quite different. However, semantics of the image contents along the ``road" can still be maintained. Row 3 and 4 show the translation from Cityscapes to KITTI and vice versa. Since both of these two datasets are collected from the cities of Germany. Most of the images share the similar structure layouts. Good results can be generated in these two experimental settings. Row 5 and 6 show the results of translating Cityscapes to its foggy version and KITTI to VKITTI-Rainy. In these experiments, robustness of the proposed method is verified to deal with different weathers (\emph{e.g.} foggy or rainy) of the city. Since the structure of a city almost remains the same, domain bias brought by different weathers mainly comes from the color and texture. In this case, quite good results can be generated by the image translation module.

\paragraph{Analysis of the inference performance}
The proposed method aims to adapt the source images into the style of the target domain. Then the translated images can be used to train the detector as augmented data. During the inference, the image translation module can be simply removed. Therefore,  the proposed method would remain the same inference time with Faster R-CNN. Similar to \cite{yuhua}, short side of the input image is scaled to 500 pixels. Using NVIDIA Tesla M40 GPU, the corresponding inference time with VGG16 is 165 ms.

\section{Conclusions and future work}
A new unsupervised domain adaptation method is proposed in this paper to solve the object detection problem in the field of autonomous driving. Extensive experiments are implemented to certify the efficacy of the proposed method. We can achieve better performance than current SOTA work through conducting adaptations both in image pixel and feature spaces. In the future, we will try to solve more complex cross-domain vision tasks such as instance segmentation or depth estimation based on the proposed method.
 
\bibliographystyle{elsarticle-num}
\bibliography{ref}

\end{document}